\documentclass{bioinfo}
\usepackage{color}
\usepackage{multirow}
\usepackage{lipsum}
\usepackage{graphicx}		
\usepackage{epstopdf}
\usepackage{comment}
\usepackage{subfig}

\copyrightyear{2014}
\pubyear{2014}

\begin{document}

\title[Mining Functionally Related Genes with Semi-Supervised Learning]{Mining Functionally Related Genes with Semi-Supervised Learning}
\author[K. Shen \textit{et~al.}]{Kaiyu Shen\,$^{1,2}$ \footnote{to whom correspondence should be addressed}, Razvan Bunescu\,$^{3}$ and Sarah E. Wyatt\,$^{1,2}$}
\address{$^{1}$Molecular and Cellular Biology, Ohio University\\
$^{2}$Environmental and Plant Biology, Ohio University\\
$^{3}$School of Electrical Engineering and Computer Science, Ohio University
}

\history{}

\editor{}

\maketitle

\begin{abstract}
\section{Motivation:}
The study of biological processes can greatly benefit from tools that automatically predict gene functions or directly cluster genes based on shared functionality. Existing data mining methods predict protein functionality by exploiting data obtained from high-throughput experiments or meta-scale information from public databases. Most existing prediction tools are targeted at predicting protein functions that are described in the gene ontology (GO). However, in many cases biologists wish to discover functionally related genes for which GO terms are inadequate. In this paper, we introduce a rich set of features and use them in conjunction with semi-supervised learning approaches in order to expand an initial set of seed genes to a larger cluster of functionally related genes.
\section{Results:} Among all the semi-supervised methods that were evaluated, the framework of learning with positive and unlabeled examples (LPU) is shown to be especially appropriate for mining functionally related genes. When evaluated on experimentally validated benchmark data, the LPU approaches\footnote{available at http://ace.cs.ohio.edu/{\tt\~{}}razvan/code/lpu-gene-mining.tar.gz} significantly outperform a standard supervised learning algorithm as well as an established state-of-the-art method. Given an initial set of seed genes, our best performing approach could be used to mine functionally related genes in a wide range of organisms.
\section{Contact:} ks280007@ohio.edu
\end{abstract}

\section{Introduction}
A common question for biologists is how to find genes or proteins that are relevant for a specific process. The question has different answers depending on the interpretation of the word ``relevant". Genes could be deemed relevant to a process because they share gene ontology (GO) terms with other genes known to be relevant. They could be involved in the same pathway, or they could directly interact.  Many \textit{in silico} approaches have been proposed to find functionally related components based on the principle of ``guilt-by-association". Functional relationships are inferred based on similarity measures that are defined in terms of biological features such as sequence alignment, secondary and tertiary structure comparisons \citep{Shen2007,Pitre2008}, gene co-expression profiles \citep{Wren2009}, phylogenetic information for non-model organisms \citep{Gaudet2011}, or topological structures of the protein-protein interactome (PPI) \citep{Przulj2004}. These features are utilized individually, but also integrated into collections of heterogeneous features for protein function prediction \citep{Costello2009,Fortney2010}. In essence, the ``guilt-by-association" strategy incrementally expands the functional group by adding genes found to be similar to the known genes.

According to an assessment by \citet{Pena-Castillo2008}, machine learning methods can successfully assign precise GO terms to genes with unknown functions. One such method is GeneMANIA \citep{Mostafavi2008a}, which uses multiple kernel learning with linear regression and was shown to perform competitively in most of the evaluations. More recently, in a similar critical assessment, 54 methods were tested on their predictions of functions for unlabeled genes \citep{Radivojac2013a}. The results of this study demonstrated again the robustness of machine learning methods. Such methods produce useful results in terms of assigning GO terms to the unknown genes. However, in many cases biologists wish to find functionally related genes for which GO terms are known to be inadequate. Biological systems have been shown to constitute scale-free networks \citep{Roy2012}, rendering GO terms inappropriate for inferring complete functional relationships. For example, gravitropism, a fundamental biological process, has been studied since Darwin and aspects of it still remain a mystery \citep{Wyatt2013}. Only three GO terms are annotated as related to gravitropism and only a third of the experimentally identified ``gravity genes'' were annotated with these terms. Over 77\% of the GO terms that were assigned to the ``gravity genes'' were shared in fewer than 10\% of those genes. This example calls into question the accuracy of using GO terms to measure ``functionally relatedness''. 

Given the success of machine learning methods in classification tasks in bioinformatics, we decided to apply learning algorithms for the problem of finding genes relevant to a particular biological process. Among the existing methods for learning classifiers, the Support Vector Machines (SVM) algorithm is a popular choice, especially for gene regulatory networks and function prediction \citep{bhardwaj2010genome,mordelet2013supervised}. The application of SVMs based on heterogeneous network combination has also been explored for almost a decade. \cite{lanckriet2004kernel} first introduced a semidefinite program approach along with the SVM (\textit{SDP/SVM}) in order to find optimal weights for a set of kernels. \citet{tsuda2005fast} suggested an SVM algorithm with linear time complexity and further indicated a condition where the kernels with optimization of weights do not always outperform one kernel with fixed weight. \citet{zhao2008gene} applied a recursive procedure for expanding negative sets in combination with one-class SVM and two-class SVMs for function prediction.

The input for out problem is a small set of {\it seed} genes known to be relevant to a biological process. The output is a larger set of genes that should be relevant for the same biological process. If the seed genes are construed as positive examples and any other gene is viewed as an unlabeled example, the problem can be cast as a task in semi-supervised learning. In our exploration of semi-supervised approaches, we have found that the framework of learning with positive and unlabeled examples (LPU) is best suited for this problem. Methods for learning from positive and unlabeled data are usually modifications of standard learning algorithms, such as SVMs \citep{liu:icdm03,Elkan:2008} and Na\"ive Bayes \citep{Denis02,Calvo2007}, or use more sophisticated models such as dynamic empirical Bayes \citep{Djebbari2008}. 

We performed a thorough evaluation of a series of machine learning approaches to the problem of growing an initial set of seed genes into a large set of genes relevant to a biological process. We evaluated a standard two-class SVM, a transductive SVM, the Laplacian SVM, a label propagation algorithm, as well as two LPU methods: the biased SVM formulation of \citet{liu:icdm03} and the weighted samples formulation of \citet{Elkan:2008}. Furthermore, we engineered a rich collection of heterogeneous features and employed feature selection in order to improve performance. We have also evaluated different strategies for selecting unlabeled genes for training, based on their relationships with the known set of seed genes. When evaluated on the \textit{Arabidopsis thaliana} benchmark dataset, the best performing system is an LPU method (Weighted SVM) trained using feature selection and unlabeled example selection. This method obtains a ranking of genes that has on average over 75\% accuracy for the top 5\% of unlabeled ranked genes. Furthermore, our approach is shown to obtain results that substantially outperform GeneMania, a state-of-the-art method for protein function prediction. Overall, we see our method as a general tool that can be utilized with various organisms and biological processes for which small sets of experimentally validated relevant genes are available.

\section{Learning Methods}
\label{sec:methods}

A straightforward approach to learning a predictor for functionally related genes uses the seed genes as positive examples and the unlabeled genes as noisy, negative examples. We used the {\it SVM}$^{light}$ implementation of binary SVMs \citep{Joachims2002} as a baseline learning method in this supervised setting.



\subsection{Transductive SVM}
Since the test examples are known in our setting, we also experimented with the {\it SVM}$^{light}$ implementation of Transductive SVMs (TSVM) \citep{joachims1999transductive}. The training dataset was created in the same way as for the binary SVM above.

\subsection{Laplacian SVM}

When plenty of unlabeled data is available, semi-supervised learning methods such as Laplacian SVMs \citep{melacci2011primallapsvm} can effectively use both the labeled and unlabeled examples to force the output of the decision function to be smooth in the intrinsic geometry inferred from the data. This means that nearby points as well as points on the same manifold tend to have the same label. This manifold regularization approach has led to state-of-the-art performance in semi-supervised settings.

\subsection{Label Propagation}
\cite{Zhou04learningwith} introduced a label propagation algorithm in which every point in the dataset iteratively spreads its label information to its neighbors until convergence to a global labeling is achieved. Like the Laplacian SVM above, the algorithm seeks to enforce a prior assumption of smoothness with respect to the intrinsic geometry of the labeled and unlabeled points.


\subsection{Learning with Positive and Unlabeled Examples}
\label{sec:LPU}

If the seed genes are construed as positive examples and any other gene is viewed as an unlabeled example, then training a binary classifier to distinguish between positive and negative genes can be seen as a problem of learning with positive and unlabeled examples (LPU). In this framework, the training data consists only of positive examples $x \in P$ and unlabeled examples $x \in U$. Following the notation of \cite{Elkan:2008}, $s(x) = 1$ if the example is positive and $s(x) = -1$ if the example is unlabeled. The true label of an example is $y(x) = 1$ if the example is positive and $y(x) = -1$ is the example is negative. Thus, $x \in P \Rightarrow s(x) = y(x) = 1$ and $x \in U \Rightarrow s(x) = -1$ i.e., the true label $y(x)$ of an unlabeled example is unknown. Given that the LPU framework obtains the best results in our problem setting (Section~\ref{sec:evaluation}), we describe in more detail our implemented versions of two state-of-the-art LPU approaches: the {\it Biased SVM}, and the {\it Weighted SVM}.

\subsubsection{Biased SVM}
In the {\it Biased SVM} formulation \citep{lee:icml03,liu:icdm03}, all unlabeled examples are considered to be negative and the decision function $f(x) = \mathbf{w}^T\phi(x) + b$ is learned using the standard soft-margin SVM formulation shown in below:
\[
\begin{array}{rl}
  \mbox{minimize:} & \dfrac{1}{2}\mathbf{w}^2 + C_P \displaystyle\sum_{x \in P}{\xi_x} + C_U \displaystyle\sum_{x \in U}{\xi_x} \\[2em]
  \mbox{subject to:} & s(x) \left(\mathbf{w}^T\phi(x) + b\right) \geq 1 - \xi_x, \hspace{0.5em} \forall x \in P \cup U \\[0.25em]
  & \xi_x \ge 0
\end{array}
\]
The capacity parameters $C_P$ and $C_U$ control how much we penalize errors on positive examples vs. errors on unlabeled examples. To find the best capacity parameters to use during training, the Biased SVM approach runs a grid search on a separate development dataset. This search is aimed at finding values for the parameters $C_P$ and $C_U$ that maximize $pr$, the product between precision $p = p(y=1|f=1)$ and recall $r = p(f=1|y=1)$. \citet{lee:icml03} show that maximizing the $pr$ criterion is equivalent with maximizing the objective $r^2/p(f=1)$, where both $r = p(f=1|y=1)$ and $p(f=1)$ can be estimated using the trained decision function $f(x)$ on the tuning dataset. We implemented the Biased SVM approach on top of the binary SVM$^{light}$ package, in which the $C_P$ and $C_U$ parameters of the Biased SVM were tuned through the $c$ and $j$ parameters ($c = C_U$ and $j = C_P/C_U$). Whereas the naive SVM tunes these parameters to optimize the accuracy with respect to the noisy label $s(x)$, the Biased SVM tunes the $c$, $j$ parameters to maximize an estimate of F1 score  with respect to the true label $y(x)$.  When training with the Gaussian kernel, the kernel width is also introduced in the grid search optimization.

\subsubsection{Weighted SVM}
In the {\it Weighted SVM} approach to LPU, \cite{Elkan:2008} make the assumption that labeled examples $\{x | s(x) = 1\}$ are selected at random from the positive examples $\{x|y(x) = 1\}$ i.e., $p(s=1|x,y=1) = p(s=1|y=1)$. Their best performing approach uses the positive and unlabeled examples to train two distinct classifiers. First, the dataset $P \cup U$ is split into a training set and a validation set, and a classifier $g(x)$ is trained on the labeling $s$ to approximate the label distribution i.e. $g(x) = p(s=1|x)$. The validation set is then used to estimate $p(s=1|y=1)$ as follows:
\begin{small}
\begin{equation}
  p(s\!=\!1|y\!=\!1) = p(s\!=\!1|x,y\!=\!1) = \dfrac{1}{|P|}\sum_{x \in P} g(x) \label{eq:prob1}
\end{equation}
\end{small}
The second and final classifier $f(x)$ is trained on a dataset of weighted examples that are sampled from the original training set as follows: 1) Each positive example $x \in P$ is copied as a positive example in the new training set with weight $p(y=1|x, s=1) = 1$; 2) Each unlabeled example $x \in U$ is duplicated into two training examples in the new dataset: a positive example with weight $p(y=1|x, s=-1)$ and a negative example with weight $p(y=-1|x, s=-1) = 1 - p(y=1|x, s=-1)$. \cite{Elkan:2008} show that the weights above can be derived as:
\begin{small}
\begin{equation}
  p(y\!=\!1|x, s\!=\!-1) = \dfrac{1\!\!-\!\!p(s\!=\!1|y\!=\!1)}{p(s\!=\!1|y\!=\!1)} \!\times\! \dfrac{p(s\!=\!1|x)}{1\!\!-\!\!p(s\!=\!1|x)} \label{eq:prob2}
\end{equation}
\end{small}
The output of the first classifier $g(x)$ is used to approximate the probability $p(s=1|x)$, whereas $p(s=1|y=1)$ is estimated using Equation~\ref{eq:prob1}. The two classifiers $g$ and $f$ are trained using SVMs. Platt scaling is used with the first classifier to obtain the probability estimates $g(x) = p(s=1|x)$, which are then converted into weights following Equations~\ref{eq:prob1} and~\ref{eq:prob2}, and used during the training of the second classifier.


\section{Data Sources and Feature Engineering}
\label{sec:features}

The semi-supervised methods require a feature vector representation of the input data. A wide range of sources were used to derive informative features, as described in the following sections.

\subsection{Protein Protein Interactions}
Protein-protein interactions (PPI) denote physical bindings between proteins, which indicate a shared functional relationships between the two corresponding genes. PPIs are typically discovered and validated through biochemical assays, affinity precipitation, yeast two-hybrid systems, or {\it in silico} predictions. Online databases such as The Arabidopsis Research Information (TAIR) \citep{Rhee2003}, InAct \citep{Kerrien2012}, BioGRID \citep{Stark2011}, and PAIR \citep{Lin2011a} contain thousands of known PPIs. We extracted only the experimentally validated interactions from these databases and used them to create a global protein-protein interaction network in which nodes stand for proteins and edges connect interacting proteins. There are in total 2322, 10680, and 15926 interactions extracted from databases with timestamps of 2010, 2011 and 2012, respectively. Given an arbitrary protein $p$ and a seed protein $s$, we compute the shortest path between $p$ and $s$ in the PPI network. The PPI feature $\phi_{PPI}(p,s)$ is then defined to have the maximum value of 1.0 when there is a direct interaction between the two proteins i.e., the shortest path contains only one edge. When the shortest path contains two or more edges, the PPI feature is calculated by subtracting 0.1 for each additional edge on the path. If $len(p,s)$ is the length of the shortest path, this means that $\phi_{PPI}(p,s) = max\{0, 1.1 - 0.1*len(p,s)\}$. Given an input protein $p$, the corresponding vector of PPI features consists of all PPI features between this protein and all seed proteins in $S$ i.e., $\Phi_{PPI}(p) = \left\{ \phi_{PPI}(p,s) | s \in S\right\}$.


\subsection{Ortholog PPI Projections}

Orthologs are genes in two species that originated from the same ancestry in the evolution tree. Ortholog genes quite often, although not always, share the same or similar functions.  Thus, an interaction between two genes in one species may be used to infer with high confidence an interaction between the orthologs in another related species. Ortholog information was retrieved from OrthoMCL \citep{li2003orthomcl} and InParanoid \citep{ostlund2010inparanoid}. A total of 14,583 PPIs from the ortholog species (\textit{S.cerevisiae}, \textit{C.elegans}, \textit{D. melanogaster}, and \textit{Z.mays}) were projected to the Arabidopsis genome. 
Using the same methodology as with the original PPI features introduced earlier in this section, we create the interaction networks for the four organisms and compute the corresponding PPI features. Using the ortholog mapping, we project these PPI features onto their counterparts in Arabidopsis. The resulting feature vectors are aggregated into an overall ortholog PPI feature vector $\Phi_{OL}(p)$.

\subsection{Biomedical Literature}

Large amounts of published biomedical articles provide information on a multitude of topics spanning many decades, much of which is stored electronically in PubMed. The automatic text processing of the PubMed articles can be used to yield a wealth of information on genes and their interactions. For each protein in the evaluation dataset, all sentences that mention that protein were automatically extracted, together with the PubMed identifiers and the publication date. The resulting 2,923,734 sentences were used to generate three types of features.

\subsubsection{Information Retrieval} For each gene $p$ in our evaluation dataset $D$, we first extracted all possible aliases and systematic names from the NCBI database. We then used the Textpresso search engine of \citet{Muller2004} to extract all sentences in PubMed that contained the gene name or any of its synonyms and created a text document $T(p)$. For every word $w$ in the vocabulary $V$ that appears in this document, we compute the standard $tf.idf$ feature weight that is normally used in classical Information Retrieval (IR) \citep{baeza-yates:book99} as $\phi_{IR}(p,w) = tf(w, p) * idf(w, p)$. The term frequency factor $tf(w,p)$ represents the number of times word $w$ appears in the extracted text $T(p)$, whereas the inverse document frequency $idf(w, p)$ is computed based on the number of different genes in the evaluation dataset $D$ for which the extracted text contains the word $w$, as shown below.
\begin{equation}
  idf(w, p) = log \frac{|D|}{\left|\left\{ p \in D | w \in T(p)  \right\}\right|} \nonumber
\end{equation}
Given an input protein $p$, the corresponding vector of IR features consists of the IR features computed for all words in the vocabulary $V$ i.e., $\Phi_{IR}(p) = \left\{ \phi_{IR}(p,w) | w \in V\right\}$. Since the evaluation dataset contains only hundreds of protein examples, the high dimensionality of the IR feature vector, when compounded with the large number of features assembled from the other sources, could negatively affect the reliability of the estimates for the learning model parameters. To alleviate this problem, we use the {\it classifier stacking} technique as follows: the seed genes are used as positive examples, while a large set of 'negative' examples is sampled at random from the Arabidopsis genome. An SVM classifier was then trained on this noisy dataset, using the high-dimensional IR feature vectors. For any given protein $p$, the margin value output by the SVM classifier is used to create a new singleton IR feature $\phi_{IR}(p)$.

\subsubsection{Shared Articles} For every gene $p$ in the dataset, we also created a set $PID(p)$ containing the PubMed identifiers of all articles found by Textpresso to contain the gene name or any of its synonyms. Correspondingly, given an arbitrary protein $p$ and a seed protein $s$, a new feature was introduced to quantify the size of the overlap between the two PMID sets i.e., $\phi_{PID}(p,s) = \left| PID(p) \cap PID(s) \right|$. Given an input protein $p$, the individual PMID feature values for all seed proteins are aggregated into a vector of PID features $\Phi_{PID}(p) = \left\{ \phi_{PID}(p,s) | s \in S \right\}$.

\subsubsection{Automatic Relation Extraction} For each input protein $p$ and any given seed $s$, we further identified the PubMed sentences that mentioned both $p$ and $s$. We then ran the relation extraction (RE) system developed by \citet{Bunescu} on all extracted sentences. The RE system was trained to identify sentences that mention interactions between proteins. For each sentence, the system outputs a probabilistic score indicating the likelihood that the sentence specifies an interaction between the two proteins. We defined a new relation extraction feature $\phi_{RE}(p,s)$ as the average of this probabilistic score, computed over all sentences containing the two proteins. Finally, given an input protein $p$, the individual RE feature values for all seed proteins are aggregated into a vector of RE features $\Phi_{RE}(p) = \left\{ \phi_{RE}(p,s) | s \in S \right\}$.

\subsection{Co-expression Profiles}

Genes with highly correlated expression profiles tend to encode interacting proteins. The most widely used technique for computing expression profiles is based on microarray data. Therefore, the expression profiles of the Arabidopsis genes in our experiments were computed based on the microarray data from TAIR \citep{Rhee2003} and the Nottingham Arabidopsis Stock Center (NASC). These two databases include 703 distinct experiments covering a total of 21975 genes. Given two arbitrary proteins $p$ and $s$, we compute the Pearson correlation coefficient (PCC) between the two genes based on their expression profiles. We then compute the rank $rank(p,s)$ of this PCC score among the set of all PCC scores between the protein $p$ and every gene in the database. The symmetric rank $rank(s,p)$ is computed in a similar way. The coexpression features $\phi_{CO}(p,s)$ are computed based on the {\it mutual rank} formula originally introduced by \cite{Obayashi2007} i.e., $\phi_{CO}(p,s) = \sqrt{rank(p,s) * rank(s,p)}$. As before, given an input protein $p$, the individual coexpression feature values for all seed proteins are aggregated into an overall vector of coexpression features $\Phi_{CO}(p) = \left\{ \phi_{CO}(p,s) | s \in S \right\}$.


\subsection{Shared Annotations}

GO provides a standard representation of proteins in terms of their associated biological processes, molecular functions, and cellular components. A GO evidence code describes the type of analysis upon which a GO term is associated with a gene. For our experiments, the GO evidence codes were restricted to the following high confidence association codes: EXP (inferred from experiment), IDA (inferred from direct assay), and IPI (inferred from physical interaction). For each protein $p \in D$ in the evaluation dataset, we extracted the set $GO(p)$ of all GO terms associated with it, restricted to the three codes above. For a protein pair $(p,s)$, the set of shared GO annotations is computed as $GO(p,s) = GO(p) \cap GO(s)$. If  $N$ is the total number of genes in the evaluation genome and $N(g)$ is the number of genes that are associated with a particular GO term $g$, then a new feature $\phi_{GO}(p,s)$ is computed as follows:
\begin{equation}
  \phi_{GO}(p,s) = log\frac{\displaystyle N}{\displaystyle min \{N(g)|g\in GO(p,s)\}} \nonumber
\end{equation}
When $GO(p,s)$ is empty, $\phi_{GO}(p,s)$ is defined to be 0. Given an input protein $p$, the individual shared GO annotation features for all seed proteins are aggregated into an overall vector of GO features $\Phi_{GO}(p) = \left\{ \phi_{GO}(p,s) | s \in S \right\}$.

Kyoto Encyclopedia of Genes and Genomes (KEGG) and AraCyc \citep{mueller2003aracyc} are two other databases of protein annotations that can be used to compute shared annotation features. KEGG provides annotations for molecular interactions and reaction pathways, whereas AraCyc is a database of metabolic pathways specifically for Arabidopsis (393 pathways and 5520 enzymes). We use a similar procedure to compute vectors of shared annotations in KEGG and AraCyc i.e., $\Phi_{KEGG}(p)$ and $\Phi_{Ara}(p)$.

\subsection{Transcription Factors and Binding Sites}

Transcription factors (TF) control the transcription levels of downstream genes. Genes that encode the same TFs are likely to share the same functionality or be involved in the same pathways. Consequently, we used the set $TF$ of transcription factors in AGRIS \citep{Palaniswamy2006} to define binary features $\phi_{TF}(p, tf)$ that are set to 1 whenever protein $p$ encodes a transcription factor $tf \in TF$. For a given protein $p$, the resulting features are aggregated into an overall TF feature vector $\Phi_{TF}(p) = \{\phi_{TF}(p, tf) | tf \in TF\}$.

Similarly, genes that share the same TF binding sites (TFBS) are more likely to be involved in the same pathway or interaction network. Therefore, we used the set $TFBS$ of transcription factor binding sites in Athamap \citep{Bulow2006} to define binary features $\phi_{TFBS}(p, tfbs)$ that are set to 1 whenever gene $p$ contains a TF binding site $tfbs \in TFBS$. For a given protein $p$, the resulting features are aggregated into an overall TFBS feature vector $\Phi_{TFBS}(p) = \{\phi_{TFBS}(p, tfbs) | tfbs \in TFBS\}$.


\subsection{Overall Feature Vector}

For any given protein $p$, the features described above were aggregated as:
\begin{small}
\begin{eqnarray*}
  \Phi(p) = [\Phi_{PPI}(p) | \Phi_{OL}(p) | \phi_{IR}(p) | \Phi_{PID}(p) | \Phi_{RE}(p) | \Phi_{CO}(p) | \\ \Phi_{GO}(p) | \Phi_{KEGG}(p) | \Phi_{Ara}(p) | \Phi_{TF}(p) | \Phi_{TFBS}(p)]^T \nonumber
\end{eqnarray*}
\end{small}
All features were normalized to lie in the [0, 1] interval. The discriminative power of each type of features was assessed individually on the seeds and unlabeled examples, with statistically significant results for all feature types.


\subsection{Feature Selection}
\label{sec:selection}

Although all types of features passed the statistical significance test, each feature type corresponds to many individual features of which only a subset is likely to be relevant for the mining task. Therefore, we performed feature selection based on a simple iterative method proposed by \cite{weston2003use} in which the selection of features was modeled through the learning of a sparse vector of feature weights as follows:
\begin{enumerate}
  \item Add a ridge term to the diagonal of the kernel matrix, to allow for non-separable data.
  \item Set $\mathbf{z} = [1,...,1]^T$, where $z_{i}$ is the weight for feature $\phi_i$. 
  \item Let $\mathbf{w}$ be the solution to the linearly separable SVM optimization problem using weighted feature vectors $\mathbf{z} * \phi(x)$.
  \item Set $\mathbf{z} \leftarrow \mathbf{z} * \mathbf{w}$ and go back to step 3 until convergence.
  \item Change feature vectors to $\phi(x) \leftarrow \mathbf{z} * \phi(x)$.
\end{enumerate}

\section{Benchmark Dataset}
\label{sec:dataset}

The benchmark data was created from the Arabidopsis Plant-Pathogen Immune Network (PPIN) \citep{Mukhtar2011} .  This gene network provides an excellent candidate for experimental evaluation because the genes can be grouped based on their involvement in the same biological process. We see this dataset as a prototypical example of a network of genes whose connectivity is driven by shared functionality.  To replicate a scenario in which a biologist wants to find more genes starting from a small set of seed genes, we first randomly selected 30, 50, or 100 positive genes $P$ from the PPIN network to serve as positive examples. We then selected 720 genes to serve as unlabeled examples $U$. The resulting dataset $P \cup U$ was split into a {\it training set} (2/3) and a {\it development set} (1/3). The {\it test set} was created from 300 positive examples (genes in the Arabidopsis PPIN) and 420 negative examples (genes not in the Arabidopsis PPIN). The learning models were trained on the positive and unlabeled examples from $P \cup U$ and then evaluated on the test dataset.

\subsection{Unlabeled Gene Selection}
\label{sec:unlabeled}

Since it is unfeasible to use all the genes in the Arabidopsis genome for training, the set $U$ of unlabeled examples was created using one of the three selection methods described below.

\subsubsection{Random}
The most straightforward method, used here as a baseline, is to randomly select genes from the Arabidopsis genome to use as unlabeled examples, including  genes that are already known to be in the PPIN network.

\begin{figure}[!t]
\centering
\includegraphics[width=\linewidth]{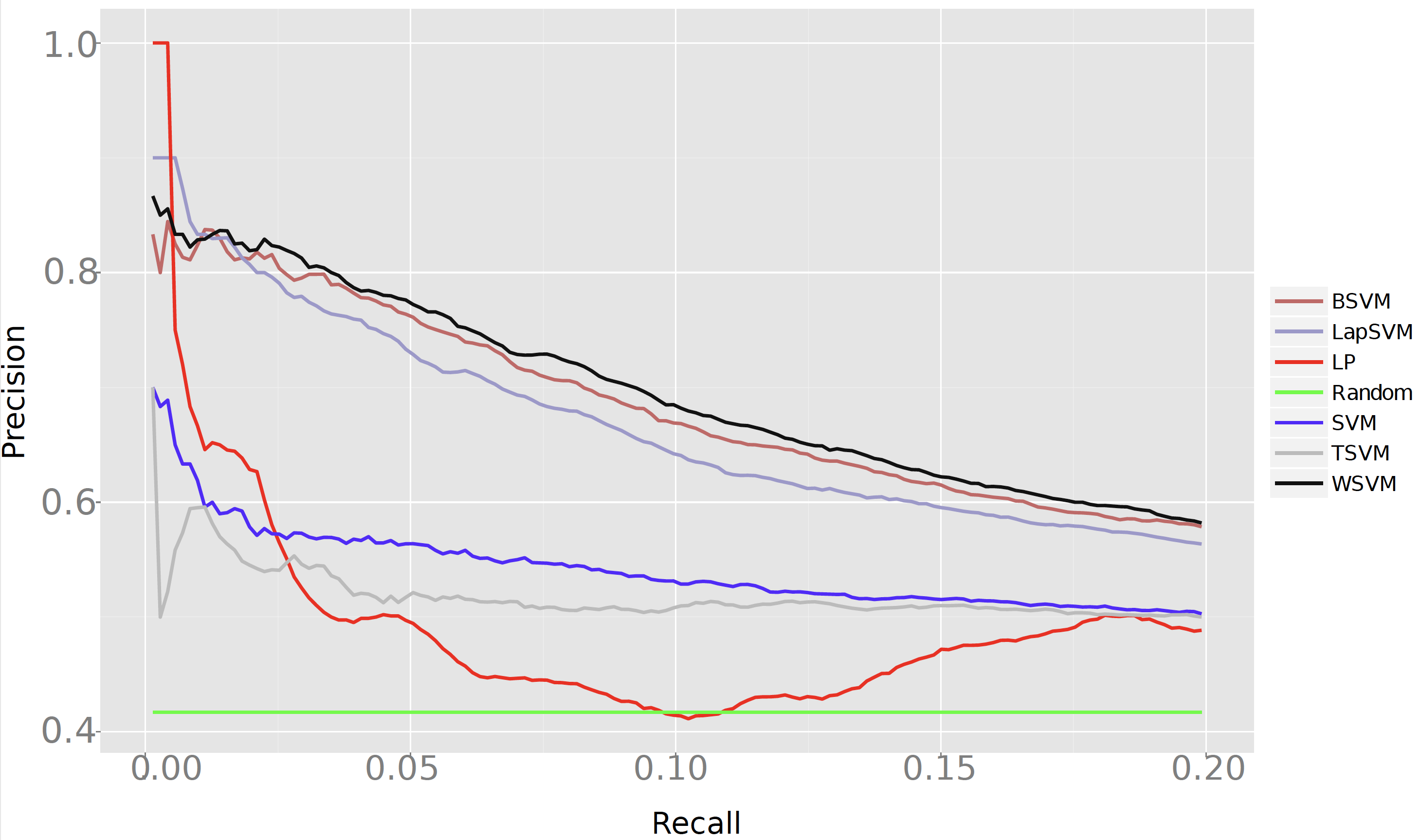}
\caption{The PR curves over the top 20\% extractions for the 6 learning methods, using data from 2012 and 30 positive seeds.} 
\label{fig:comparison1}
\end{figure}

\subsubsection{GO Similarity}

We also introduced two approaches for unlabeled gene selection informed by GO terms. Let $T(g)$ be the set of GO terms associated with a gene $g$, and $T(P) = \displaystyle \cup_{g \in P} T(g)$ the set of GO terms associated with the positive set of genes $P$. Furthermore, let $sim(t_i,t_j)$ by the similarity of two GO terms $t_i$ and $t_j$, as computed by the G-SESAME method \citep{Du2009}. We define the similarity of a gene $g$ to the positive set $P$ as:
\begin{equation}
  sim(g, P) = \sum_{t_i \in T(g)} \sum_{t_j \in T(P)} sim(t_i,t_j)
\end{equation}
The genes in the entire Arabidopsis genome were ranked based on their similarity $sim(g, P)$ with the positive set. The genes that had the lowest similarity were then selected as unlabeled examples.
 
\subsubsection{GO Distance} Let $d(t_j,t_i)$ be the length of the shortest path between terms $t_i$ and $t_j$ in the GO graph. Let also $d$ and $D$ be some predefined minimum and maximum distance, respectively. The set of unlabeled genes was then selected to contain the genes $g$ that satisfied the following distance constraint:
\begin{equation}
  \exists t_i \! \in \! T(g)\!-\!T(P), t_j \! \in \! T(P) \; s.t. \; d \leq d(t_i,t_j) \leq D
\end{equation}
We used the constants $d=3$ and $D=8$, as they gave the best performance on the development dataset.

\section{Experimental Evaluation}
\label{sec:evaluation}

We implemented and comparatively evaluated the 6 methods described in Section~\ref{sec:methods}: the binary SVM, transductive SVM (TSVM), Laplacian SVM (LapSVM), label propagation (LP), biased SVM (BSVM), and weighted samples SVM (WSVM).  The kernel based algorithms were evaluated with both a linear and a Gaussian kernel. Results are reported on the test data, by selecting the best performing kernel on development data.  Thus, BSVM obtained the best performance with a linear kernel, whereas the other algorithms obtained the best performance with a Gaussian kernel. Fig.~\ref{fig:comparison1} shows the precision vs. recall (PR) curves of the 6 methods on the top 20\% extractions from the test data. Additionally, the green line shows the performance of a baseline that assigns binary labels at random. The feature vectors (Section~\ref{sec:features}) were created based on data available in 2012, using 30 seed genes in the positive set $P$. The results show a significant gap between the 3 top performing algorithms (BSVM, WSVM, and LapSVM) and the other 3 algorithms (binary SVM, TSVM, and LP).  WSVM with the Gaussian kernel obtains the best overall performance, whereas TSVM and LP do worse than the standard binary SVM.

\begin{figure}[!t]
\centering
\includegraphics[width=\linewidth]{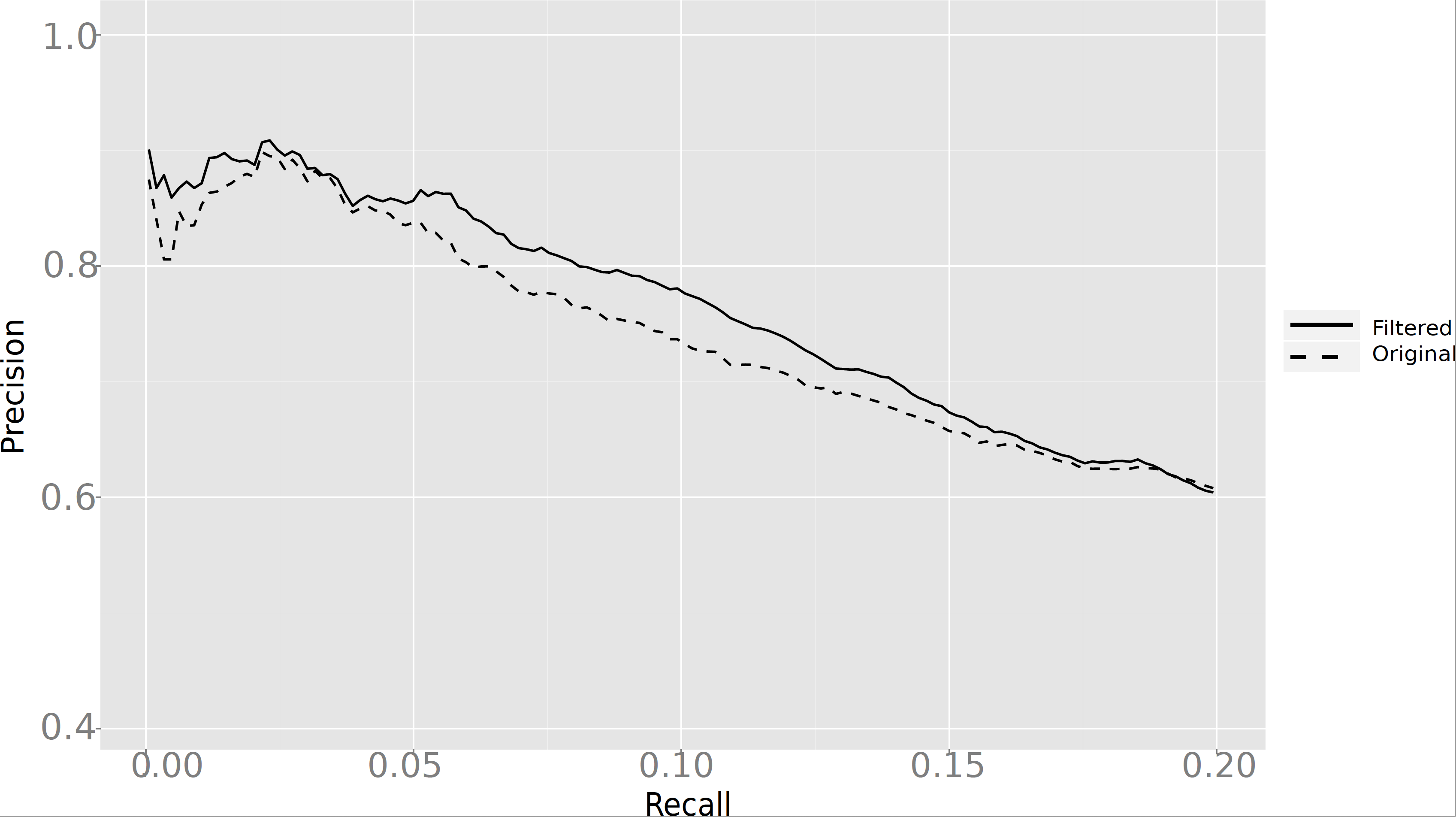}
\caption{WSVM performance with feature selection (solid line) and without feature selection (dashed line).}
\label{fig:comparison2}
\end{figure}
To assess the impact of feature selection, we selected the top performing models -- WSVM and BSVM -- and evaluated them with ({\it filtered}) and without ({\it original}) feature selection. Table~\ref{tab:originalfilter} shows the area under the PR curve for the top 20\%  extractions (AUC20) obtained by the two methods under the two  feature vector scenarios, for 30, 50, and 100 seeds genes in the positive set $P$. For both methods, feature selection was beneficial when there were at least 50 seed genes in $P$. The most substantial improvement from feature selection is observed for WSVM when using 100 positive seeds. Correspondingly, Fig.~\ref{fig:comparison2} shows the PR curve of WSVM with 100 positive seeds in training, with and without feature selection. To increase readability, both figures start at a 40\% precision level.

\begin{table}[th]
 \centering
  \caption{AUC20  for BSVM and WSVM on original and filtered feature sets.}
    \begin{tabular}{r |r r r r}
    \hline
          Seeds & \multicolumn{2}{c}{WSVM} & \multicolumn{2}{c}{BSVM}  \\
    \hline
          & Original & Filtered & Original & Filtered \\
    30 & {\bf 0.690} & 0.680	& 0.680 & 0.673 \\
    50 & 0.711 & {\bf 0.723} & 0.706 & 0.716 \\
    100 & 0.733 & {\bf 0.758} & 0.739 & 0.748 \\
    \hline
    \end{tabular}
  \label{tab:originalfilter}
\end{table}


In order to estimate the impact of the amount of available information, we ran multiple experiments in which the extraction of the input features was constrained to content that appeared before a predetermined year: 2010, 2011, and 2012. 
Table~\ref{tab:timestamp} shows the AUC20 performance of the two top performing models, for each of the three years, using linear and Gaussian kernels. The results show that WSVM with Gaussian kernel is still the best performing model. While, as expected, data available from more recent years leads to better results, the performance for year 2010 is still close to the best performance for 2012.

\begin{table}[th]
  \centering
  \caption{AUC20 for WSVM and BSVM, constrained by years.}
    \begin{tabular}{r|rrrr}
    \hline
          Algorithm & \multicolumn{2}{c}{WSVM} & \multicolumn{2}{c}{BSVM}  \\
    \hline
          & Gaussian & linear & Gaussian & linear \\
    2010 & {\bf 0.655} & 0.605	& 0.535 & 0.635 \\
    2011 & {\bf 0.675} & 0.605	& 0.605 & 0.645 \\
    2012 & {\bf 0.690} & 0.650 & 0.615 & 0.680 \\
    \hline
    \end{tabular}
  \label{tab:timestamp}
\end{table}


Table~\ref{tab:negatives} shows the AUC values of the best performing model WSVM, when using the three methods for unlabeled example selection described in Section~\ref{sec:unlabeled}. Both GO-based methods outperform the random sampling of unlabeled examples, with GO Similarity being the best method overall.

\begin{table}[th]
  \centering
  \caption{AUC20 for WSVM for the 3 unlabeled example selection methods.}
    \begin{tabular}{r|ccc}
    \hline
     Seeds  & Random & GO Similarity & GO Distance \\
    \hline
    30 & 0.690 & {\bf 0.717} & 0.712 \\
    50 & 0.711 & {\bf 0.725} & 0.723 \\
    100 & 0.738 & {\bf 0.741} & {\bf 0.741} \\
    \hline
    \end{tabular}
  \label{tab:negatives}
\end{table}




\subsection{Comparison with GeneMANIA}

We further compared our WSVM-based approach with GeneMANIA, a state-of-the-art method developed by \citet{Mostafavi2008a}. Of all feature types described in Section~\ref{sec:features}, GeneMANIA uses PPI, protein domain, physical interactions, co-localization, pathway and co-expression data to construct functional association networks. A composite functional association network is then built as a weighted average of the individual association networks. Genes that are relevant for a particular biological process are then predicted through label propagation on the composite network, starting from a set of seed genes.
We provided both WSVM and GeneMania with the same 30, 50 and 100 positive samples. For each number of seed genes, we randomly sampled the seed genes and repeated the experiment 30 times. We report the performance in terms of micro-averaged \textit{Precision@N}, where N is the number of genes returned by GeneMANIA. Since the extracted genes may repeat in the 30 random trials, we also report \textit{Precision@N} by counting only the number of unique true positive genes extracted across all 30 trials. For WSVM we report performance when using all features from Section~\ref{sec:features} (WSVM$_2$) and also when using only the same features as GeneMANIA (WSVM$_1$). The results are listed in Table~\ref{tab:addlabel} and show that WSVM outperforms GeneMANIA substantially in all settings, with best performance obtained when using all the features. A one-tailed t-test of statistical significance for the comparisons between WSVM and GeneMANIA results in p-values less than 0.001, with the exception of the last column, where the p-value is less than 0.05. When analyzed in the context of the other experimental results, this significant difference in performance is not surprising. GeneMANIA uses label propagation, which was shown above (Fig. ~\ref{fig:comparison1}) to substantially underperform the LPU methods.

\begin{table*}[t]
  \centering
  \caption{Weighted samples SVM (WSVM) vs. GeneMANIA (GM), in terms of Precision@N, where N = \# genes extracted by GeneMANIA.}
    \begin{tabular}{c|cccccc}
    \hline
          Model & \multicolumn{2}{c}{30 seeds} & \multicolumn{2}{c}{50 seeds} & \multicolumn{2}{c}{100 seeds} \\
    \hline
          & Redundant & Unique & Redundant & Unique & Redundant & Unique \\
    GM & 21.0\% (637/3030)  & 	12.9\% (224/1742)  & 17.0\% (117/690) & 11.5\% (40/348) & 29.4\% (236/803)  & 31.7\% (127/400) \\
    WSVM$_1$ & 54.7\% (1658/3030) & 56.7\% (472/833) & 62.3\% (430/690) & 52.4\% (204/389) & 65.9\% (529/803) & 59.4\% (173/291) \\
    WSVM$_2$ & {\bf 65.3\%} (1980/3030) & {\bf 58.7\%} (287/489) & {\bf 69.6\%} (480/690) & {\bf 53.4\%} (226/423) & {\bf 77.1\%} (619/803) & {\bf 60.5\%} (201/332) \\
    \hline
    \end{tabular}
  \label{tab:addlabel}
\end{table*}

\section{Conclusion}

A common problem encountered by researchers in their study of biological processes  is that of expanding an initial set of seed genes to a larger cluster of functionally related genes, for which GO terms are often inadequate. We approach this task as a semi-supervised learning problem that is solved automatically by exploiting features derived from a rich set of sources such as biomedical literature and gene expression data. Of the six learning models that were evaluated in this paper, the best performance was obtained by the two models that were trained in the LPU framework. In particular, the Weighted SVM was shown to significantly outperform an existing state-of-the-art method on the Arabidopsis PPIN benchmark. Experimental evaluation on data from different years shows that the performance of the LPU methods is only slightly impacted by the removal of data from more recent years. Furthermore, the models were shown to benefit from feature selection and a ranking of unlabeled examples informed by GO relations.

\section{Acknowledgments}
We would like to thank the Ohio Supercomputer Center for providing computing time for the data analysis and the Ohio University Genomics Facility for providing assistance with the microarray experiments.

\paragraph{Funding\textcolon}
Kaiyu Shen and Sarah Wyatt were supported by National Science Foundation grant IOS-1147087.

\bibliographystyle{bioinformatics}

\bibliography{manuscript}
\end{document}